\documentclass{article}
\usepackage{ijcai24}

\usepackage{times}
\usepackage{soul}
\usepackage{url}
\usepackage[hidelinks]{hyperref}
\usepackage[utf8]{inputenc}
\usepackage[small]{caption}
\usepackage{graphicx}
\usepackage{amsmath}
\usepackage{amsthm}
\usepackage{booktabs}
\usepackage{algorithm}
\usepackage{amsfonts}
\usepackage{amsthm}
\usepackage{algorithmic}
\usepackage{multirow}
\usepackage[switch]{lineno}
\theoremstyle{definition}
\newtheorem{theorem}{Theorem}
\newtheorem{lemma}[theorem]{Lemma}
\usepackage{wrapfig}
\usepackage{makecell}
\usepackage{multirow}
\usepackage{pdflscape}
\usepackage{varwidth}
\usepackage[figuresright]{rotating}

\title{WeatherGNN: Exploiting Meteo- and Spatial-Dependencies for Local Numerical Weather Prediction Bias-Correction}

\author{
Binqing Wu$^{1,2}$\and
Weiqi Chen$^{1}$\and
Wengwei Wang$^1$\and
Bingqing Peng$^1$ \and
Liang Sun$^{1}$\And
Ling Chen$^{2}$\\
\affiliations
$^1$ DAMO Academy, Alibaba Group\\
$^2$ College of Computer Science and Technology, Zhejiang University\\
\emails
\{binqingwu, lingchen\}@cs.zju.edu.cn,\\
\{jarvus.cwq, duoluo.www, pengbingqing.pbq, liang.sun\}@alibaba-inc.com
}

\begin{document}

\maketitle
\begin{abstract}
Due to insufficient local area information, numerical weather prediction (NWP) may yield biases for specific areas. Previous studies correct biases mainly by employing handcrafted features or applying data-driven methods intuitively, overlooking the complicated dependencies between weather factors and between areas. To address this issue, we propose WeatherGNN, a local NWP bias-correction method that utilizes Graph Neural Networks (GNNs) to exploit meteorological dependencies and spatial dependencies under the guidance of domain knowledge. Specifically, we introduce a factor GNN to capture area-specific meteorological dependencies adaptively based on spatial heterogeneity and a fast hierarchical GNN to capture dynamic spatial dependencies efficiently guided by Tobler's first and second laws of geography. Our experimental results on two real-world datasets demonstrate that WeatherGNN achieves the state-of-the-art performance, outperforming the best baseline with an average of 4.75 \% on RMSE.
\end{abstract}

\section{Introduction}
Numerical weather prediction (NWP) has become the widely accepted and effective method for weather forecasting~\cite{Bauer2015Quiet}, developing from solving mathematical equations under physical laws~\cite{weather_textbook} to incorporating deep learning methods~\cite{bi2023accurate,chen2023fuxi,nguyen2023climax}. Despite its advancements, due to insufficient local area information~\cite{gneiting2005weather,yoshikane2022bias}, NWP may be still biased for specific areas. This kind of deviation leads to significant discrepancies in downstream applications, e.g., wind power forecasting~\cite{han2022short}.

\begin{figure}
    \centering
    \includegraphics[width=0.48\textwidth]{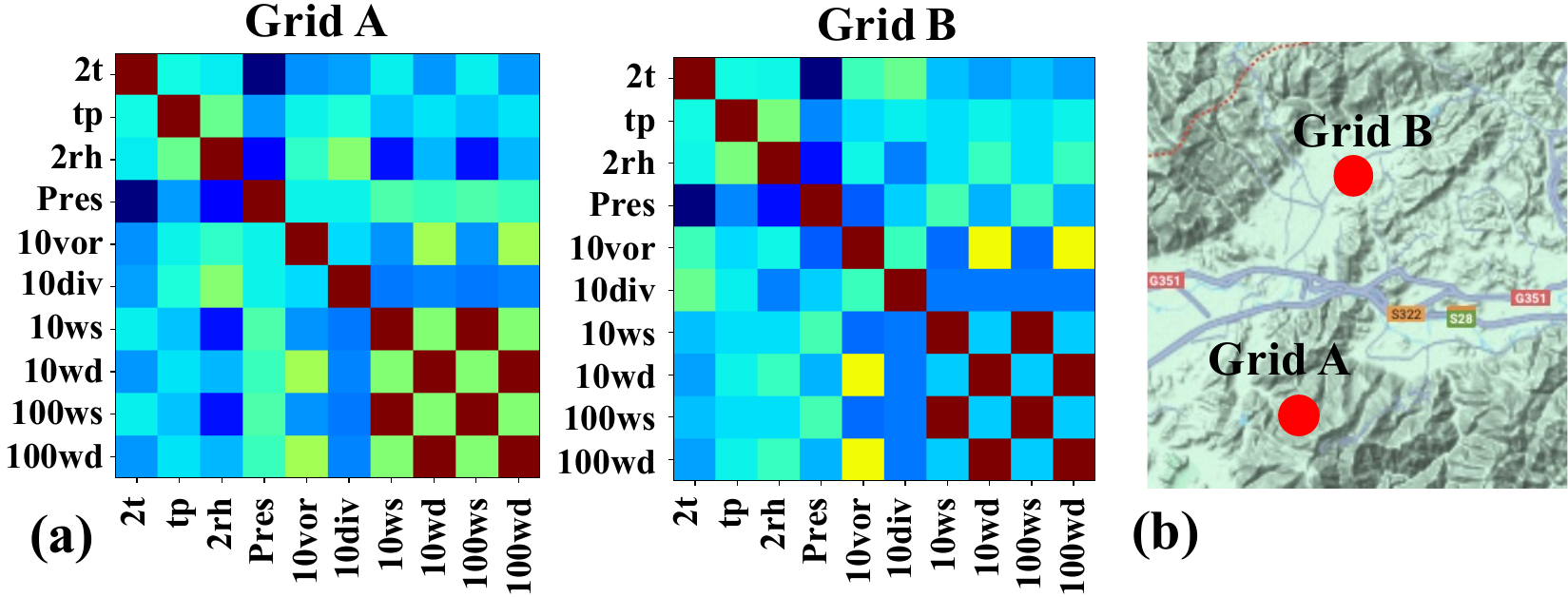}
    \caption{(a) Pearson Correlation Matrices between Weather Factors at Different Grids. (b) Locations and Terrains of Two Grids. Grid A is on an uphill, while B is in a valley. The distance between Grid A and B is around 18 kilometers.}
    \label{fig:pearson}
\end{figure}

Local NWP bias-correction aims to correct biases of NWP data for specific areas, given additional local area information, e.g., weather factor observations and terrain \cite{zhang2023deep}. Early local NWP bias-correction methods usually apply statistical rules and shallow machine learning~\cite{delle2006ozone,cui2012bias,durai2014evaluation}, which cannot model the complex non-linearity within NWP. Recently, some deep learning methods have achieved impressive success owing to their strong representative abilities. They mainly rely on Recurrent Neural Networks (RNNs)~\cite{li2022numerical,yang2022correcting}, Convolutional Neural Networks (CNNs)~\cite{han2021deep}, or their hybrid methods~\cite{han2022short}. Despite the promising results of existing methods, we argue that two domain-specific problems are still overlooked.

First, these methods mainly correct weather factors independently, rarely considering the complicated dependencies between them. In reality, weather factors interact with each other and these dependencies are strongly correlated with heterogeneous geographical characteristics, e.g., terrain. Fig. \ref{fig:pearson} illustrates an example, where each grid represents a specific area. Grid A, positioned uphill, and Grid B, located in a valley, exhibit distinct correlation patterns even if they are close.

\begin{figure}
    \centering
    \includegraphics[width=0.28\textwidth]{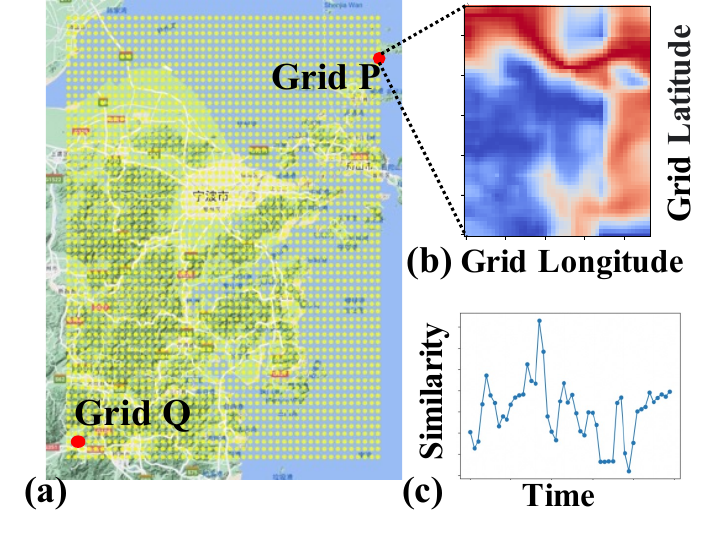}
    \caption{(a) Grid Distribution and Terrain of Ningbo Dataset. (b) DTW Similarity of Half-month 100m Wind Speed between Grid P and Other Grids. (c) DTW Similarity of 100m Wind Speed with a Six-hour Time Range between Grid P and Grid Q.}
    \label{fig:geo_dtw}
\end{figure}

Second, previous methods often neglect or only concentrate on grid-based dependencies between areas, failing to capture complicated dependencies between areas. Due to geographical impacts and atmospheric motions, the weather conditions of one area have a grid-agnostic and dynamic influence on those of other areas. Fig. \ref{fig:geo_dtw} illustrates an example of 100m wind speed. Over a half-month duration, the similarity of wind speed between Grid P and other areas closely aligns with the terrain similarity, as depicted in Fig. \ref{fig:geo_dtw}(b), showcasing grid-agnostic dependencies. Meanwhile, the similarity of wind speed between Grid P and Grid Q, computed within a six-hour time range of that duration, undergoes temporal evolution, as revealed in Fig. \ref{fig:geo_dtw}(c).

To this end, we propose WeatherGNN, a GNN-based method that exploits meteorological dependencies and spatial dependencies for local NWP bias-correction. WeatherGNN adopts two GNNs to capture area-specific meteorological dependencies adaptively and dynamic spatial dependencies efficiently, guided by geography principles, i.e., spatial heterogeneity and Tobler’s first and second laws. To the best of our knowledge, WeatherGNN is the first work that incorporates GNNs and domain knowledge for this task. The main contributions are as follows:

\begin{itemize}
    \item Introduce a factor GNN that combines factor embeddings and geographical embeddings derived from heterogeneous geographical characteristics to construct factor graphs for each area, which can learn area-specific meteorological dependencies adaptively.
    \item Introduce a fast hierarchical GNN that constructs a static hierarchy based on pre-defined geographical and meteorological distances and adjusts the hierarchy according to NWP data, which can capture dynamic spatial dependencies. Notably, we introduce a fast hierarchical message passing module, inspired by Tobler’s laws, that exchanges fine-grained messages among grids with stronger dependencies and coarse-grained messages through repeated operations among grids with weaker dependencies, which can achieve linear complexity concerning the number of grids.
    \item Experimental results on two real-world datasets demonstrate the superiority of WeatherGNN, outperforming the best baselines with an average of 4.75 \% on RMSE.
\end{itemize}

\section{Related Work}\label{sec:related:work}
\textbf{Local NWP Bias-correction.} Early local bias-correction methods evolve from statistical methods, e.g., correcting maximum and minimum values \cite{durai2014evaluation} or the variance and average~\cite{zhang2019bias}, to shallow machine learning methods, e.g., applying Kalman filter \cite{delle2006ozone} and deep belief networks~\cite{HU2021120185}, which cannot capture non-linear dependencies within NWP. Recently, some deep learning methods have been applied due to their strong representative abilities. They mainly rely on RNNs~\cite{li2022numerical,yang2022correcting}, CNNs~\cite{han2021deep}, or their hybrid methods~\cite{han2022short,zhang2023deep}. Some advanced methods have also been applied to local NWP bias-correction. For example, Dl-Corrector-Remapper~\cite{ge2022dl} uses Adaptive Fourier Neural Operators (AFNOs) to learn continuous status of the atmosphere and correct four weather factors. CM2Mc-LPJmL~\cite{hess2022physically} utilizes Generative Adversarial Networks (GANs) to improve distributions of the precipitation output. However, since they mainly correct one or a few weather factors independently, they rarely consider complicated dependencies between weather factors. In addition, existing methods barely consider or only capture gird-based dependencies between areas via CNNs, which cannot capture gird-agnostic dependencies between areas.\\
\noindent\textbf{State-of-the-art Deep Learning Methods for Weather Applications.} 
SOTA methods have shown promising results in many weather applications. 
As discussed above, for local NWP bias-correction, RNNs and CNNs have achieved great improvement. 
For precipitation prediction, CNNs, GANs, and vision transformers show impressive ability \cite{ravuri2021skilful,zhang2023skilful,gao2022earthformer}.
For weather downscaling, CNNs \cite{hu2019runet} and Physics-informed neural networks \cite{esmaeilzadeh2020meshfreeflownet} have garnered advancements.
For medium-range global weather forecasting, methods ranging from AFNOs~\cite{pathak2022fourcastnet} to hierarchical transformers~\cite{chen2023fengwu,chen2023fuxi,nguyen2023climax} have demonstrated significant success ~\cite{rasp2023weatherbench}. Recently, GraphCast~\cite{lam2023learning} designs a multi-scale unweighted mesh graph for global forecasts, which exhibits the strong potential of GNNs to model atmospheric dynamics. However, these methods only focus on simple spatial dependencies and barely consider dependencies between weather factors explicitly.\\
\noindent\textbf{Graph Neural Networks.} GNN is a general framework for constructing neural networks on graph-structured data. The most common form of a GNN is the message-passing paradigm \cite{wu2020comprehensive,zhou2020graph}. During each message-passing layer, the representations of a node are updated according to messages aggregated from the graph neighborhood of that node. The update and aggregate operators are implemented by neural networks. Due to the strong representative ability on graphs, GNNs have achieved impressive success in many real-world applications \cite{wu2022graph}, ranging from graph classification \cite{ma2023multi}, traffic forecasting \cite{jiang2023megacrn}, and air quality estimation \cite{chen2023group}. However, GNNs face the challenge of designing graphs for specific applications and executing efficient message passing across graphs. Exploring methods to harness the advantages of GNNs and customize them for local NWP bias-correction is still an ongoing challenge.

\section{WeatherGNN}
We formalize local NWP bias-correction and describe how to model area-specific meteorological dependencies and dynamic spatial dependencies using WeatherGNN.

\subsection{Problem Definition}
The goal of local NWP bias-correction is to correct biases of NWP data for specific areas, given additional local area information. Formally, NWP data for areas are represented as a sequence $\left[ \mathcal{X}_{i} \right]_{i=1}^T \in \mathbb{R}^{N \times F_{\text{e}} \times T}$, where $N$ denotes the number of grids \footnote{Areas tend to appear as a rectangle containing $H \times W$ grids, and here we flatten the shape with $N = H \times W$.}. $F_{\text{e}}$ denotes the number of weather factors, e.g., temperature, humidity, and wind speed, and $T$ is the length of the sequence. Similarly, $\left[ \mathcal{Y}_{i} \right]_{i=1}^T \in \mathbb{R} ^ {N \times F_{\text{e}} \times T}$ represents the corresponding weather factor observations. Moreover, each grid has associated geographical information denoted as $\mathcal{Z} \in \mathbb{R} ^ {N \times F_{\text{g}}}$, where $F_{\text{g}}$ is the dimension of geographic features, e.g., longitude, latitude, and altitude. The local NWP bias-correction is formulated as learning a function $f(\cdot)$ that maps NWP data to local weather factor observations incorporating geographical information:
\begin{equation}
\begin{aligned}
& \{ [ \mathcal{X}_{i} ]_{i \in \Omega(t)}; \mathcal{Z} \} \xrightarrow{f(\cdot)} \mathcal{Y}_{t}, \\
& \Omega(t) = \{t-\tau, \cdots, t-1, t, t+1, \cdots, t+\tau \},
\end{aligned}
\end{equation}
where $\Omega(t)$ denotes a temporal window around time step $t$ with length $T = 2\tau+1$ , as we consider a period of time before and after $t$ when correcting $\mathcal{X}_t$. For simplicity, in the remainder of the paper, we denote $\left[ \mathcal{X}_{i} \right]_{i \in \Omega(t)}$ and $\mathcal{Y}_{t}$ by $\boldsymbol{X}$ and $\boldsymbol{Y}$, respectively.

\subsection{Model Overview} 
\begin{figure}
    \centering
    \includegraphics[width=0.46\textwidth]{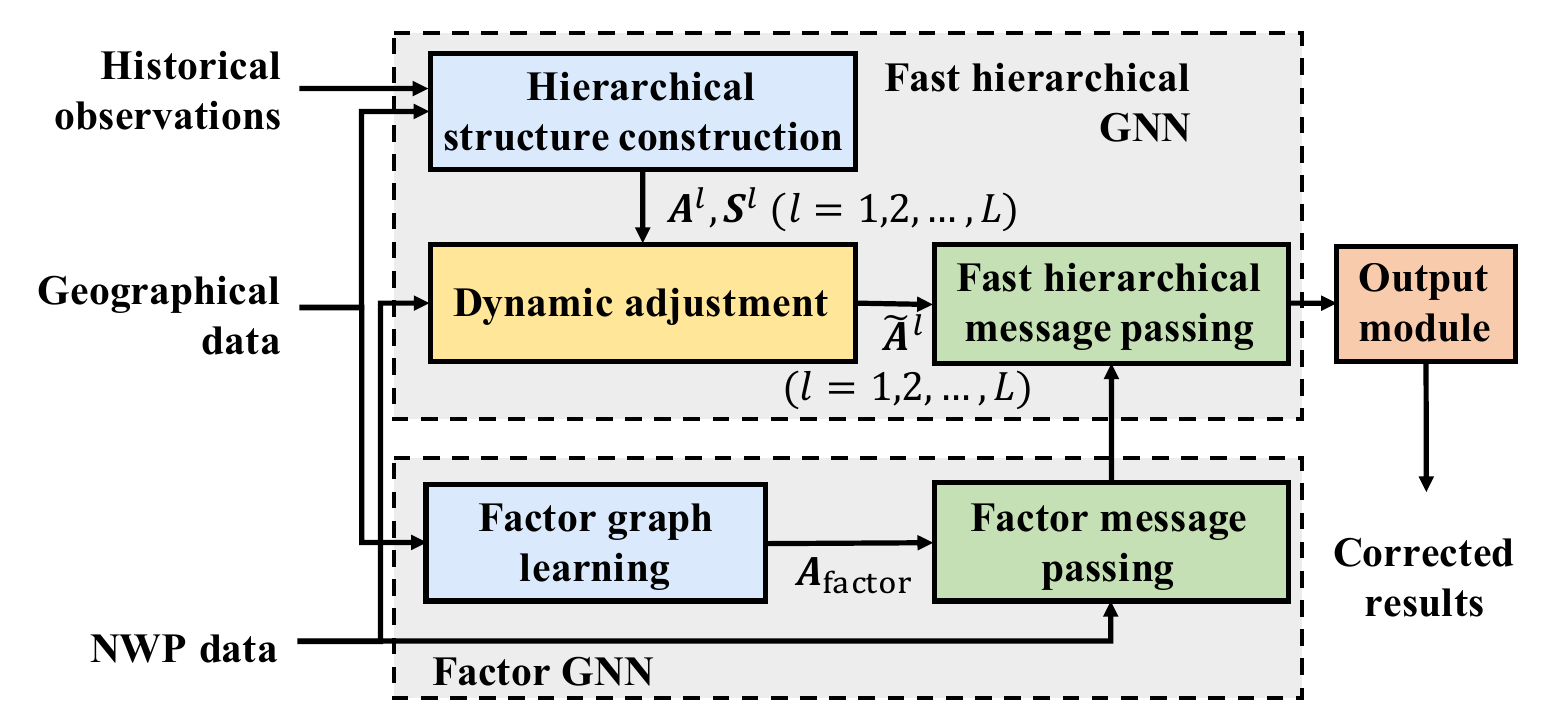}
    \caption{An Overview of WeatherGNN.}
    \label{fig:framework}
\end{figure}

As illustrated in Fig. \ref{fig:framework}, WeatherGNN adopts a two-branch architecture. A factor GNN and a fast hierarchical GNN are designed to capture area-specific meteorological dependencies and dynamic spatial dependencies, respectively. Subsequently, an output module is applied to yield corrected results.

\subsection{Learning Area-specific Meteorological Dependencies via Factor GNN}
According to \textbf{\textit{Spatial heterogeneity}} \cite{anselin2013spatial}, which states that geographic variables exhibit uncontrolled variance, meteorological dependencies, significantly influenced by geographic variables, are expected to vary based on heterogeneous geographical characteristics consequently. However, defining meteorological dependencies for each area is challenging due to issues of applicability and non-linearity. To address this problem, we introduce a factor GNN that constructs a factor graph for each area based on the geographical characteristics of each area in a data-driven way.

\noindent\textbf{Factor graph learning.} The factor graph learning (FGL) module first randomly initializes a learnable factor embedding dictionary $\boldsymbol{E}_{\text{f}} \in \mathbb{R}^{F_{\text{e}} \times d_{\text{e}}}$ for weather factors, where each row corresponds to the embedding vector for each factor, and $d_{\text{e}}$ is the dimension of factor embedding. This embedding is optimized during model training, aiming to capture the latent features associated with each factor. Then, we take the geographical information of each grid $\mathcal{Z}$ as input and encode it with an MLP to obtain the geographical embedding matrix $\boldsymbol{E}_{\text{g}} = \mathrm{MLP}(\mathcal{Z}) \in \mathbb{R}^{N \times d_{\text{e}}}$, where $N$ is the number of grids. $\boldsymbol{E}_{\text{g}}$ represents geographical characteristics of each grid.
Given the factor embedding shared by all grids and the geographical embedding of each grid, we can infer dependencies between weather factors for each grid by:
\begin{equation}
\begin{aligned}
    & \boldsymbol{E}_i = \boldsymbol{E}_{\text{f}} + \boldsymbol{E}_{\text{g},i}, \\
    & \boldsymbol{A}_{\text{f},i} = \mathrm{Softmax}(\boldsymbol{E}_i \boldsymbol{E}_i^\top), \\
    & \boldsymbol{A}_{\text{f}} = \{\boldsymbol{A}_{\text{f},i} \}_{i=1}^N,     
\end{aligned}
\end{equation}
where $\boldsymbol{E}_i \in \mathbb{R}^{F_{\text{e}} \times d_{\text{e}}}$ is the grid-specific factor embedding of grid $i$ considering geographical characteristics. $\boldsymbol{A}_{\text{f},i} \in \mathbb{R}^{F_{\text{e}} \times F_{\text{e}}}$ represents the adjacency matrix of the factor graph of grid $i$. The FGL module provides structured representations of meteorological dependencies for each grid/area, which can facilitate an understanding of how these weather factors interact, taking into account heterogeneous geographical characteristics.

\noindent\textbf{Factor Message Passing.} Enhanced by FGL, the factor message passing for a single layer can be formulated as:
\begin{equation}\label{equ:fgnn}
    \boldsymbol{H}_{i} = (\boldsymbol{I}_\text{f} + \boldsymbol{A}_{\text{f},i}) \boldsymbol{X}_{i} \boldsymbol{W}_{\text{f},i} + \boldsymbol{b}_{\text{f},i}
\end{equation}
where $\boldsymbol{X}_{i}\in \mathbb{R}^{F_{\text{e}} \times T}$ is the NWP data of grid $i$, and $\boldsymbol{H}_{i} \in \mathbb{R}^{F_{\text{e}} \times d}$ is the corresponding output considering meteorological dependencies. $\boldsymbol{I}_\text{f}$ is the identity matrix. $\boldsymbol{W}_{\text{f},i} \in \mathbb{R}^{F_{\text{e}} \times T \times d}$ and $\boldsymbol{b}_{\text{f},i} \in \mathbb{R}^{F_{\text{e}} \times d} $ denote the learnable weights and bias, respectively. They can be generated by two smaller parameter matrices and grid-specific factor embeddings, i.e., $\boldsymbol{W}_{\text{f},i} = \boldsymbol{E}_i \boldsymbol{W}_{\text{w}}$ and $\boldsymbol{b}_{\text{f},i} = \boldsymbol{E}_i \boldsymbol{W}_{\text{b}}$. 
Note that $\boldsymbol{W}_{\text{w}} \in \mathbb{R}^{d_{\text{e}} \times T \times d}$ and $\boldsymbol{W}_{\text{b}}  \in \mathbb{R}^{d_{\text{e}} \times d}$ are shared by all grids.
$\boldsymbol{W}_{\text{f},i}$ and $\boldsymbol{b}_{\text{f},i}$ can be interpreted as capturing grid-specific patterns guided by a set of shared patterns discovered from all grids. This factor GNN is performed for each grid in parallel, obtaining the final output $\boldsymbol{H} = \{\boldsymbol{H}_{i} \}_{i=1}^N \in \mathbb{R}^{N \times F_{\text{e}} \times d}$. According to Lemma \ref{lemma:complexity}, the factor GNN has an $O(N)$ complexity.

\subsection{Learning Dynamic Spatial Dependencies via Fast Hierarchical GNN}
Due to the complex terrains and varying weather conditions, capturing dependencies between areas is challenging. To address this issue, we propose a Fast Hierarchical GNN (FHGNN) that can effectively capture dynamic spatial dependencies by combining geographical distances, historical weather similarity, and NWP data.
Specifically, we first calculate the initial spatial proximity of grids by considering geographical distances and historical weather similarity. Then, we construct the hierarchical structure through an iterative process of clustering grids into multi-level clusters (super-grids). In this structure, grids within the same cluster at lower levels exhibit stronger dependencies and are expected to interact more closely with each other. Following this, we optimize spatial proximity by leveraging an attention matrix derived from NWP data, considering current weather conditions. Subsequently, we introduce a Fast Hierarchical Message Passing (FHMP) module to capture dynamic spatial dependencies efficiently. This module facilitates fine-grained interactions between grids at a lower level and coarse-grained interactions at a higher level. Notably, FHGNN demonstrates a linear complexity concerning the number of grids, in contrast to the quadratic complexity of a vanilla GNN.

\noindent\textbf{Constructing static hierarchical structure based on pre-defined geographical and meteorological distances.} According to \textbf{\textit{Tobler's first law of geography}} \cite{miller2004tobler}, which states that everything is related to everything else, but near things are more related than distant things, we first determine the neighboring grids for each grid. While the Euclidean distance is an intuitive measure of spatial proximity, it falls short of capturing the diverse surface features. 

Since variations in altitude significantly influence hydrology, vegetation, and other geographical phenomena, we consider altitude as a crucial factor reflecting geographical features and the Earth's surface shape, in addition to latitude and longitude. We compute the 3D geographical distance matrix $\boldsymbol{D}$ using latitude, longitude, and altitude data obtained from geographical information $\mathcal{Z}$. In addition, as illustrated in Fig. \ref{fig:geo_dtw}, historical weather similarity can also indicate terrain similarity. Thus, we calculate the meteorological distance matrix $\boldsymbol{W}$ by employing DTW~\cite{muller2007dynamic} to measure the distance between weather sequences for each grid. The two distance matrices are calculated as follows:
\begin{equation}
\boldsymbol{D}_{ij} = \sqrt{\lambda_{\text{lat}}d_{\text{lat},ij}^2+\lambda_{\text{lon}}d_{\text{lon},ij}^2+\lambda_{\text{alt}}d_{\text{alt},ij}^2},
\end{equation}
\begin{equation}
\begin{aligned}
    \boldsymbol{W}_{ij}^f = \mathrm{DTW}(\boldsymbol{\mathsf{Y}}_{i}^f,\boldsymbol{\mathsf{Y}}_{j}^f), \boldsymbol{W}_{ij} = \frac{1}{F_{\text{e}}} \Sigma_{f=1}^{F_{\text{e}}}\boldsymbol{W}_{ij}^f,    
\end{aligned}
\end{equation}
where $d_{\text{lat},ij}$, $d_{\text{lon},ij}$, $d_{\text{alt},ij}$, are the latitude, longitude, and altitude distance between grid $i$ and $j$, respectively, with $\lambda_{(\cdot)}$ being the corresponding coefficients to adjust the importance of these distances for calculating the geographical distance.
$\boldsymbol{\mathsf{Y}}_{i}^f$ and $\boldsymbol{\mathsf{Y}}_{j}^f$ are the observation time series of the $f$-th weather factor of grid $i$ and $j$ in the whole training data, respectively. $\boldsymbol{W}_{ij}^f$ and $\boldsymbol{W}_{ij}$ are the DTW distance matrix of grid $i$ and $j$ for weather factor $f$ and for all weather factors, respectively.

Gaussian kernels are then employed to calculate two initial adjacency matrices, and these matrices are combined to form the spatial adjacency matrix. The process is formulated as:
\begin{equation} \label{inter_a}
\begin{aligned}
& {\boldsymbol{A}}_{\text{geo}}  = \text{exp} (-\boldsymbol{D}^2/\sigma_{\text{geo}}^2),\quad
{\boldsymbol{A}}_{\text{met}}  = \text{exp} (-\boldsymbol{W}^2/\sigma_{\text{met}}^2), \\
& \boldsymbol{A}_{\text{spatial}} = {\boldsymbol{A}}_{\text{geo}} + {\boldsymbol{A}}_{\text{met}},
\end{aligned}
\end{equation}
where $\sigma_{(\cdot)}$ are hyperparameters to control the scale of the corresponding Gaussian kernels, which ensures both matrices have a similar range of values.

Since grids have different neighbors at different spatial scales, we adopt a clustering algorithm, e.g., K-means, to cluster grids into multi-level super-grids based on $ \boldsymbol{A}_{\text{spatial}}$. This process forms hierarchical graphs, offering insights into the neighbors of each grid at different spatial scales. The clustering procedure is iterated $L$ times. At the $l$-th iteration, grids at level $l$ are clustered into the next coarsened super-grids at level $l+1$. The assignment matrix from this clustering, denoted as $\boldsymbol{S}^l \in \{0,1\}^{N^l \times N^{l+1}}$, signifies the assignment of each grid at level $l$ to a super-grid at level $l+1$, with each row containing a single 1 and the rest as 0s. Here, $N^l$ represents the number of grids at level $l$. In addition, we compute the spatial adjacency matrix $\boldsymbol{A}^l$ for each level. The construction procedure of the hierarchical structure can be formulated as:
\begin{equation}
\begin{aligned}
    & \boldsymbol{S}^l \leftarrow \text{K-means}(\boldsymbol{A}^l), \\
    & \boldsymbol{A}^{l} = (\boldsymbol{S}^{l-1})^\top \boldsymbol{A}^{l-1} \boldsymbol{S}^{l-1}, \quad
    \boldsymbol{A}^{0} = \boldsymbol{A}_{\text{spatial}},    
\end{aligned}
\end{equation}
where the spatial adjacency matrix $\boldsymbol{A}^{l} \in \mathbb{R}^{N^{l} \times N^{l}}$ ($l = 1,2,\cdots,L$) signifies a coarsened spatial graph that captures the connectivity strength between super-grids at level $l$. 

Moreover, we introduce the mask matrix $\boldsymbol{M}^l$ to make $\boldsymbol{A}^{l}$ sparse. The $(i, j)$-th entry $\boldsymbol{M}^l_{ij}$ is set to 1 if grids $i$ and $j$ belong to the same cluster at level $l+1$, and 0 otherwise. This assignment is based on the observation that grids within the same cluster exhibit stronger spatial dependencies, making them more desirable to retain. Consequently, $\boldsymbol{M}^l$ reflects the most crucial neighboring grids at level $l$.

\noindent\textbf{Dynamic adjustment of hierarchical structure based on NWP data.} Spatial dependencies are further influenced by dynamic weather conditions. For example, on a windy day, a grid has stronger dependencies with its upwind neighboring grids compared to a windless day. Thus, we encode the input NWP data $\boldsymbol{X}$ and utilize the attention matrix to adaptively adjust the spatial adjacency matrices. Inspired by attention mechanisms \cite{vaswani2017attention}, this dynamic adjustment is formulated as \footnote{We introduce the aggregation and sparsification procedure in a dense matrix form for ease of understanding, while in our implementation, the cluster assignment, mask, and spatial adjacency are represented by a sparse matrix form.}:
\begin{equation} \label{equ:h}
\begin{split}
    &\boldsymbol{H}^0_{\text{Q}} = \boldsymbol{X} \boldsymbol{W}_{\text{Q}}, \quad
    \boldsymbol{H}^0_{\text{K}} = \boldsymbol{X} \boldsymbol{W}_{\text{K}},\\
    &\boldsymbol{H}^{l}_{\text{Q}} = (\boldsymbol{S}^{l-1})^\top   \boldsymbol{H}^{l-1}_{\text{Q}}, \quad
    \boldsymbol{H}^{l}_{\text{K}} = (\boldsymbol{S}^{l-1})^\top   \boldsymbol{H}^{l-1}_{\text{K}}, \\
\end{split}
\end{equation}
\begin{equation} \label{equ:adjust}
    \tilde{\boldsymbol{A}}^{l} = \boldsymbol{M}^l  \odot \mathrm{Softmax}(\boldsymbol{H}_{\text{Q}}^{l}{\boldsymbol{H}_{\text{K}}^{l}}^\top /\sqrt{d} \odot    \boldsymbol{A}^l),
\end{equation}
where $\boldsymbol{H}^0_{\text{Q}},\boldsymbol{H}^0_{\text{K}} \in \mathbb{R}^{N \times d}$ are calculated by linear projections of input NWP data $\boldsymbol{X} \in \mathbb{R}^{N \times F_{\text{e}} \times T}$ with parameters $\boldsymbol{W}_{\text{Q}}, \boldsymbol{W}_{\text{K}} \in \mathbb{R}^{F_{\text{e}} \times T \times d}$, respectively. These representations are iteratively aggregated according to the cluster assignment matrix, generating representations $\boldsymbol{H}^{l}_{\text{Q}}, \boldsymbol{H}^{l}_{\text{K}} \in  \mathbb{R}^{N^l \times d}$ for super-grids at each level. $\odot$ denotes element-wise product. Different from existing attention mechanisms, the mask $\boldsymbol{M}^l $ is used to make the spatial adjacency matrix sparse, where only interactions of grids belonging to the same cluster are reserved. Note that this dynamic adjustment only modifies the spatial adjacency matrices at each level and does not change the assignment matrices between levels.

\begin{figure}
    \centering
    \includegraphics[width=0.3\textwidth]{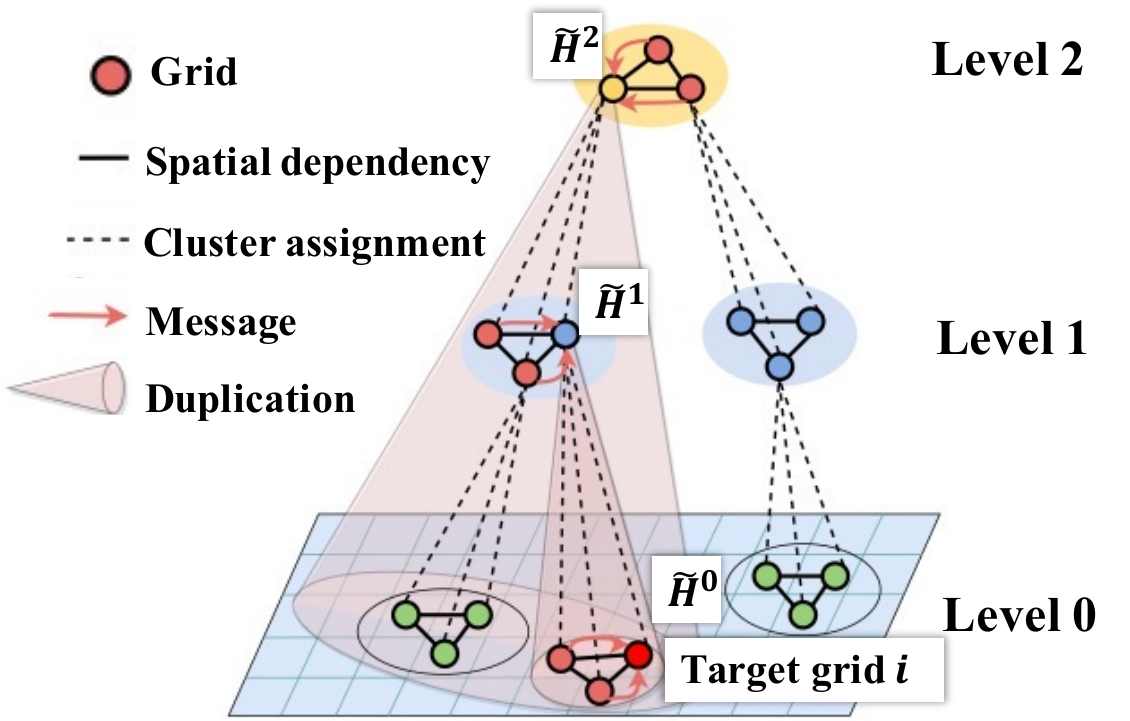}
    \caption{An Overview of Fast Hierarchical Message Passing.}
    \label{fig:fhmp}
\end{figure}

\noindent\textbf{Fast hierarchical message passing.} According to \textbf{\textit{Tobler'second law of geography}} \cite{miller2004tobler}, which states that the phenomenon external to a geographic area of interest affects what goes on inside, we correct NWP biases of one area by considering the impact from other areas. However, given that the number of grids in local NWP bias-correction is typically larger than in other common spatial-temporal applications, e.g., traffic forecasting, applying the Vallinia message-passing module with quadratic complexity on the constructed dynamic graphs poses an efficiency challenge. Therefore, we introduce the Fast Hierarchical Message Passing (FHMP) module to capture dynamic spatial dependencies efficiently. Inspired by Tobler's first and second laws of geography, the main idea of FHMP is to conduct fine-grained message passing among low-level grids which have stronger relationships but coarse-grained message passing at high levels which have weaker relationships. The process can be formulated as:
\begin{equation} \label{fhmp}
    \begin{split}
        \text{Message} &: \hat{\boldsymbol{H}}^l = \boldsymbol{H}^l \boldsymbol{W}^l, \boldsymbol{H}^{l} = (\boldsymbol{S}^{l-1})^\top \boldsymbol{H}^{l-1}, \\
        \text{Aggregation} &: \tilde{\boldsymbol{H}}^l_{v^l} = \frac{1}{|\mathcal{N}_{v^l}|} \sum_{u^l}\tilde{\boldsymbol{A}}_{u^lv^l}^l \hat{\boldsymbol{H}}^l_{u^l}, \\
        & \forall v^l \in [0,1,2, \cdots N^l-1], \mathcal{N}_{v^l} = \{u^l | \tilde{A}^l_{u^lv^l} \neq 0\}, \\
        \text{Duplication} &: \boldsymbol{H}^{\prime} = \tilde{\boldsymbol{H}}^0 + \sum_{l=1}^L \prod_{i=0}^{l-1}\boldsymbol{S}^i \tilde{\boldsymbol{H}}^l, \\
        \text{Update}&: \boldsymbol{H}^{\prime \prime} = \boldsymbol{H} + \boldsymbol{H}^{\prime}\boldsymbol{W},
        \end{split}
\end{equation} 
where $\hat{\boldsymbol{H}}^l \in \mathbb{R}^{N^l \times d}$ are calculated by linear projections of $\boldsymbol{H}^l$, representing the message of grids at level $l$, and $\boldsymbol{H}^0=\boldsymbol{H}$ is the output of the factor GNN (see Eq. \ref{equ:fgnn}). For a specific grid $v_l$ at level $l$, its neighbors $\mathcal{N}_{v^l}$ are identified by the spatial adjacency matrix $\tilde{\boldsymbol{A}}^l$. $v_l$ then aggregates its neighbors' messages according to the adjacency weights in $\tilde{\boldsymbol{A}}^l$. The duplication stage is the key design of FHMP module \footnote{Since each row of $\prod_{i=0}^{l-1}\boldsymbol{S}^i$ only contains one 1 and others are 0s, the matrix multiplication
of $\prod_{i=0}^{l-1}\boldsymbol{S}^i \tilde{\boldsymbol{H}}^l$ is implemented with indexing operation.}. It duplicates the messages from the super-grids at each level to their containing grids at level 0 based on assignment matrices $[S^{i}]^{L-1}_{i=0}$, and the grids at level 0 belonging to the same super-grid share the same message from $\hat{\boldsymbol{H}}^l$, as illustrated in Fig. \ref{fig:fhmp}. With the hierarchical structure\footnote{The conclusion holds if we establish $\lfloor \mathrm{log}_k N\rfloor + 1$ levels, where $k$ is the cluster size.}, all grid-pairs at level 0 can interact. After collecting the messages from all levels, the hidden representations of grids at level 0 are updated based on integrated messages with a residual connection. According to Lemma \ref{lemma:complexity}, FHMP in FHGNN have a linear complexity w.r.t. the number of grids. The proof is in Appendix.

\begin{lemma}\label{lemma:complexity}
The complexity of factor GNN and fast hierarchical GNN is $O(N)$, where $N$ is the number of grids.
\end{lemma}

\begin{table*}[htbp]
\centering
\resizebox{\textwidth}{!}{
\begin{tabular}{c|cccccccccccc|cc}
\hline
Dataset & Factor & Metric & NWP & BiLSTM-T & \begin{tabular}[c]{@{}c@{}}HybridCBA\end{tabular} & \begin{tabular}[c]{@{}c@{}}ConvLSTM\end{tabular} & AFNO & Swin & STGCN & HGCN & \begin{tabular}[c]{@{}c@{}}MegaCRN\end{tabular} & \begin{tabular}[c]{@{}c@{}}WeatherGNN\end{tabular} & \begin{tabular}[c]{@{}c@{}}$\Delta$Baseline\end{tabular} & \begin{tabular}[c]{@{}c@{}}$\Delta$NWP\end{tabular} \\
\hline
\multirow{10}{*}{Ningbo} & \multirow{2}{*}{100ws} & MAE & 2.26 & 1.98 & 1.53 & 1.37 & \underline{0.84} & 0.91 & 0.94 & 0.89 & 0.87 & \textbf{0.80} & 4.76\% & 64.60\% \\
 &  & RMSE & 2.74 & 2.70 & 2.34 & 2.03 & \underline{1.17} & 1.26 & 1.31 & 1.25 & 1.21 & \textbf{1.14} & 2.56\% & 58.39\% \\
 & \multirow{2}{*}{10ws} & MAE & 1.38 & 1.37 & 1.18 & 0.91 & \underline{0.58} & 0.67 & 0.65 & 0.63 & 0.61 & \textbf{0.56} & 3.45\% & 54.42\% \\
 & & RMSE & 1.71 & 1.69 & 1.65 & 1.28 & \underline{0.81} & 0.87 & 0.91 & 0.88 & 0.85 & \textbf{0.79} & 2.47\% & 53.80\% \\
 & \multirow{2}{*}{h} & MAE & 15.37 & 14.48 & 12.32 & 7.61 & 5.86 & 5.96 & 6.35 & \underline{5.81} & 5.95 & \textbf{5.63} & 3.10\% & 63.37\% \\
 & & RMSE & 18.28 & 18.44 & 15.29 & 10.35 & \underline{7.26} & 7.32 & 8.29 & 7.28 & 7.80 & \textbf{7.12} & 1.93\% & 61.05\% \\
 & \multirow{2}{*}{2t} & MAE & 1.28 & 1.20 & 1.19 & 1.16 & 0.93 & 1.10 & 1.12 & 1.04 & \underline{0.91} & \textbf{0.90} & 1.10\% & 29.69\% \\
 & & RMSE & 1.74 & 1.54 & 1.58 & 1.51 & 1.27 & 1.34 & 1.47 & 1.31 & \textbf{1.23} & \textbf{1.23} & 0.00\% & 29.31\% \\
 & \multirow{2}{*}{tp} & MAE & 0.143 & 0.412 & 0.331 & 0.309 & \underline{0.123} & 0.240 & 0.282 & 0.212 & 0.131  &  \textbf{0.115} & 6.50\% & 19.58\% \\
 & & RMSE & 0.495 & 0.811 & 0.763 & 0.655 &  \underline{0.391} & 0.587 & 0.627 & 0.502 & 0.435 &  \textbf{0.339} & 13.30\% & 31.52\% \\
\hline
\multirow{10}{*}{Ningxia} & \multirow{2}{*}{100ws(U)} & MAE & 2.93 & 2.69 & 2.33 & 2.22 & 2.00 & 2.12 & 2.33 & 2.08 & \underline{1.95} & \textbf{1.78} & 8.72\% & 39.25\% \\
 &  & RMSE & 3.83 & 3.41 & 2.99 & 2.87 & 2.58 & 2.81 & 3.01 & 2.69 & \underline{2.54} & \textbf{2.39} & 5.91\% & 37.60\% \\
& \multirow{2}{*}{100ws(V)} & MAE & 3.39 & 2.79 & 2.67 & 2.51 & \underline{2.25} & 2.30 & 2.51 & 2.29 & 2.27 & \textbf{2.13} & 5.33\% & 37.17\% \\
 &  & RMSE & 4.52 & 3.59 & 3.43 & 3.24 & \underline{2.93} & 3.11 & 3.22 & 3.03 & 2.94 & \textbf{2.69} & 8.19\% & 40.49\% \\
& \multirow{2}{*}{10ws(U)} & MAE & 1.79 & 1.65 & 1.44 & 1.40 & \underline{1.22} & \underline{1.22} & 1.52 & 1.33 & 1.23 & \textbf{1.18} & 3.28\% & 34.08\% \\
 &  & RMSE & 2.38 & 2.16 & 1.87 & 1.84 & \underline{1.60} & 1.64 & 1.97 & 1.73 & \underline{1.60} & \textbf{1.53} & 4.38\% & 35.71\% \\
& \multirow{2}{*}{10ws(V)} & MAE & 2.03 & 1.73 & 1.55 & 1.50 & \underline{1.32} & 1.41 & 1.56 & 1.38 & 1.35 & \textbf{1.21} & 8.33\% & 40.39\% \\
 &  & RMSE & 2.74 & 2.24 & 2.02 & 1.96 & \underline{1.75} & 2.08 & 2.01 & 1.85 & 1.80 & \textbf{1.59} & 9.14\% & 41.97\% \\
& \multirow{2}{*}{2t} & MAE & 2.81 & 2.71 & 2.72 & 2.44 & 2.27 & 2.27 & 2.54 & \underline{2.20} & 2.28 & \textbf{2.19} & 0.45\% & 22.06\% \\
 &  & RMSE & 3.74 & 3.49 & 3.54 & 3.15 & 2.91 & 2.89 & 3.27 & \textbf{2.79} & 2.92 & \underline{2.80} & -0.36\% & 25.13\% \\
\hline
\end{tabular}
}
\caption{Bias-correction Performance Comparison on Ningbo and Ningxia Datasets.}
\label{tab:ningbo}
\end{table*}

\begin{table}[htbp]
\centering
\scriptsize
\begin{tabular}{ll|cc|cc}
\hline
\multicolumn{2}{l|}{Factor} & \multicolumn{2}{c|}{Ningbo 100ws} & \multicolumn{2}{c}{Ningxia 100ws(U)} \\
\hline
\multicolumn{2}{l|}{Metric} & MAE & RMSE & MAE & RMSE \\
\hline
\multirow{6}{*}{Variant} & shared-F & 0.83 & 1.18 & 1.87 & 2.47 \\
& no-F & 0.92 & 1.26 & 1.94 & 2.55 \\ 
& geo-H & 0.91 & 1.26 & 2.03 & 2.60 \\
& met-H & 0.89 & 1.25 & 1.92 & 2.57 \\
& static-H & 0.85 & 1.22 & 1.89 & 2.50 \\
& no-H & 0.96 & 1.31 & 2.33 & 2.82 \\
\hline
\multicolumn{2}{l|}{WeatherGNN} &  \textbf{0.80} &  \textbf{1.14} & \textbf{1.78} & \textbf{2.39} \\
\hline
\end{tabular}
\caption{Ablation Study on 100m Wind Speed.}
\label{tab:ablation}
\end{table}

\subsection{Bias-Correction Output}
Given the output $\boldsymbol{H}^{\prime\prime} \in \mathbb{R}^{N \times d}$  of FHGNN, we feed it into an MLP-based decoder to produce corrected weather factors for each grid at the target time step at once. We adopt the L1 loss function to compare the difference between corrected results and ground truth, which can be formulated as: $\mathcal{L}(\hat{\boldsymbol{Y}}, \boldsymbol{Y}) = \sum_{f=1}^{F_{\text{e}}} \alpha_f \Vert \hat{\boldsymbol{Y}}^f - \boldsymbol{Y}^f \Vert_1$, where $ \hat{\boldsymbol{Y}} = \mathrm{MLP} \left( \boldsymbol{H}^{\prime\prime} \right)$,
and $\hat{\boldsymbol{Y}}^f$ and $\boldsymbol{Y}^f$ are the corrected results and the ground truth of the weather factor $f$. $\alpha_f$ is a hyperparameter representing the corresponding weight of weather factor $f$.

\section{Experiments}
\subsection{Datasets}
Since local NWP bias-correction requires high-precision geographical data and direct mapping between NWP and observations of multiple weather factors, current datasets cannot meet such requirements. Thus, we collect two real-world bias-correction datasets: Ningbo and Ningxia, covering two representative terrain types in China. Each grid in both datasets has three types of data, i.e., geographical data, NWP data, and weather factor observations. Geographical data contain latitude, longitude, and altitude. Altitude data are from Digital Elevation Model (DEM) data. In particular, DME data are commonly used in geographic information systems to represent terrain, whose positive and negative values can denote land and ocean, respectively. 
\textbf{Ningbo} dataset depicts a coastline area with latitude range $28.85^\circ$N$-30.56^\circ$N and longitude range $120.91^\circ$E$-122.29^\circ$E. There are 58$\times$47 grids with a grid size of 0.03 degree in latitude and longitude. 
Both NWP and observational weather data have hourly records including 10 weather factors from 1/Jan/2021 to 1/Apr/2021. 
\textbf{Ningxia} dataset features mountainous and hilly area with latitude range $34.5^\circ$N$-42^\circ$N and longitude range $106^\circ$E$-116^\circ$E. There are 31$\times$41 grids with grid size of 0.25 degree in latitude and longitude. 
Both NWP and observational weather data have hourly records including 8 weather factors from 1/Jan/2021 to 1/Jan/2022. In our experiments, we divide each dataset into training/validation/test subsets using a 7:1:2 ratio in chronological order. We define a temporal window of 7 time steps for NWP bias-correction, including the target time step, as well as the previous and next three steps.

\subsection{Baselines and Experimental Settings}
We compare WeatherGNN with three groups of methods. (1) Specific for bias-correction, including BiLSTM-T~\cite{yang2022correcting}: using BiLSTM considering multiple covariates for bias-correction; HybridCBA~\cite{han2022short}: combining CNN, BiLSTM, and attention mechanism to correct factors. (2) Vision methods used in many weather applications, including ConvLSTM~\cite{convlstm}: extending LSTM with convolutional gates; AFNO~\cite{guibas2021efficient}: adopting Fourier neural operator to capture features adaptively; Swin~\cite{liu2021swin}: constructing a hierarchical transformer by a shifted windowing scheme. (3) GNN-based methods used in spatial-temporal applications, including STGCN~\cite{stgcn}: deploying graph convolution and temporal convolution to capture spatial and temporal dependencies; HGCN~\cite{hgcn}: operating hierarchical GNNs to utilize hierarchical spatial dependencies; MegaCRN~\cite{jiang2023megacrn}: introducing adaptive graphs to learn underlying spatial dependencies. 

Baselines and WeatherGNN are implemented with Pytorch, executed on a server with one 32GB Tesla V100 GPU card, and well-tuned according to the performance on the validation set. The hyperparameter settings are summarized in Appendix. We optimize WeatherGNN using Adam optimizer with an initial learning rate of 0.003 and set the maximum number of epochs to 200. We halt training when the validation loss does not decrease for 15 consecutive epochs.

\subsection{Performance Comparison}
We use Mean Absolute Error (MAE) and Root Mean Square Error (RMSE) to measure the performance of all methods. We present the results of common weather factors for main comparisons, e.g., wind speed (ws), pressure(h), temperature at 2m (2t), and total precipitation (tp). Table \ref{tab:ningbo} shows the results, where \textbf{Bold} and \underline{Underline} indicate the best and second best performance, respectively. $\Delta$ denotes the relative improvement between WeatherGNN and best baselines/NWP. We can observe that: 
(1) NWP for specific areas can be significantly corrected, and WeatherGNN has an average improvement of over 41\% compared with the original NWP.
(2) Although baselines, especially AFNO and MegaCRN, have shown competitive performance, WeatherGNN achieves the \textbf{best performance} in 19 out of 20 cases across two datasets, outperforming the best baselines with average 4.50\% on MAE and 4.75\% on RMSE. Despite the remarkable improvement, WeatherGNN shows a slight inferiority in correcting 2t compared to the best baseline. The reason may be that the 2t factor varies slowly in time and space. The rapid changes in wind speeds may interfere with the correction of 2t due to the meteorological and spatial dependency modeling of WeatherGNN.
(3) The bias-correction of WeatherGNN on wind speed is significant. Given the fact that the mechanism of wind formation is complex and wind speed changes over time and space rapidly, it is possible that the dynamic adjustment in the fast hierarchical GNN helps WeatherGNN adapt to weather dynamics.

\subsection{Model Analysis}
\textbf{Ablation study.} To evaluate the effectiveness of the factor GNN and fast hierarchical GNN of WeatherGNN, we conduct ablation studies on correcting the 100m wind speed of two datasets. 
For factor GNN, we design variants:
(1) shared-F: only using the factor embedding to construct a shared factor graph for all grids, ignoring spatial heterogeneity;
(2) no-F: replacing the factor graphs with a one-layer MLP to encode the weather information without considering the meteorological dependencies explicitly.
For fast hierarchical GNN, we design variants:
(3) geo-H: constructing the hierarchy only using geographical distances;
(4) met-H: constructing the hierarchy only using DTW similarity of historical weather factor time series;
(5) static-H: removing the dynamic adjustment and only adopting the static hierarchy when performing fast hierarchical message passing;
(6) no-H: only using the $\boldsymbol{A}_{\text{spatial}}$ calculated by Eq. \ref{inter_a} to model spatial dependencies between grids without constructing the hierarchy.

All designs in WeatherGNN are proven to be effective based on the results in Table \ref{tab:ablation}. We can observe that:
(1) Terrain is crucial for local NWP bias-correction, even when it is solely used as input (as shown in "no-F"). By comparing "shared-F" and "no-F", we observe that modeling the dependencies between factors explicitly can improve the correction outcomes. Notably, making such dependencies adapt to the changing terrain can further enhance the correction effect, which validates the effectiveness of our factor GNN.
(2) Constructing a hierarchical structure with geographical and meteorological distances significantly improves the bias-correction performance. The meteorological distance is found to be more effective than geographical distance, possibly because the influence of geography on weather factors is partly reflected in the weather factor sequences and DTW similarity can help to discover underlying spatial dependencies. Besides, the meteorological distance is helpful to unveil remote spatial dependencies. Furthermore, the hierarchical design integrating both factors achieves the best results.

\begin{figure}
    \centering
    \includegraphics[width=0.48 \textwidth]{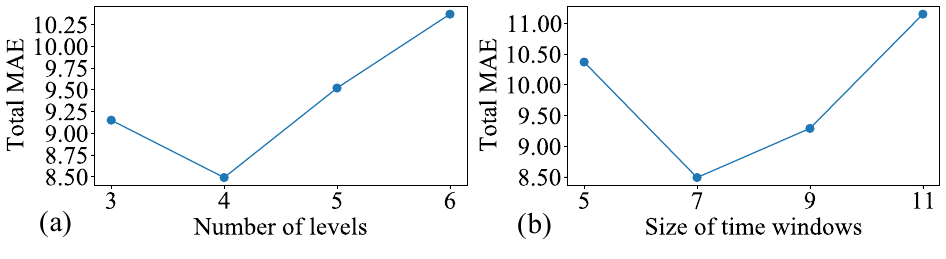}
    \caption{Results of Hyperparameter Study}
    \label{fig:hyper}
\end{figure}

\noindent\textbf{Hyperparameter study.} We adjust two hyperparameters, i.e., the number of levels (\# levels) and the size of the time window, to investigate the effect of the static hierarchy and time length. We use total MAE of all factors in Ningbo dataset as the metric. We set \# levels to 3, 4, 5, and 6. From Fig.\ref{fig:hyper} (a), we find the correction performance is optimal when \# level is 4. It is possible because smaller \# levels fail to capture comprehensive multi-scale spatial dependencies, whereas larger \# levels result in fewer grids in each super-grid, limiting the effectiveness of capturing local fine-grained dependencies. We set the size of the time window to 5, 7, 9, and 11. From Fig.\ref{fig:hyper} (b), we find the time window is essential to the correction performance, with the optimal choice being 7. The possible reason is that smaller windows lack sufficient information to capture weather changes, while larger windows introduce excessive uncertainty about weather conditions.

\begin{figure}
    \centering
    \includegraphics[width=0.45 \textwidth]{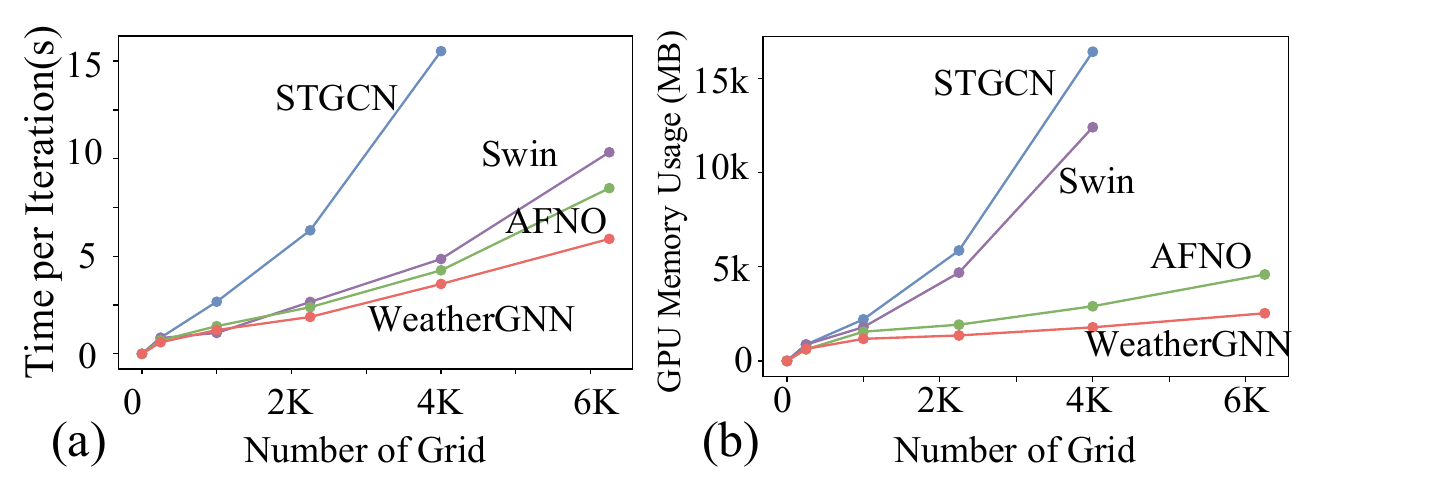}
    \caption{Model Efficiency Comparison.}
    \label{fig:eff}
\end{figure}

\noindent\textbf{Model efficiency.} 
We evaluate the efficiency of WeatherGNN and baselines by increasing the number of grids and measuring running time and GPU memory usage. Fig. \ref{fig:eff} illustrates that WeatherGNN (linear complexity) performs faster and requires less GPU memory than baselines. The efficiency advantages of WeatherGNN indicate good scalability, showing strong potential for larger regions, e.g., continental or global applications.

\begin{figure}
    \centering
    \includegraphics[width=0.3\textwidth]{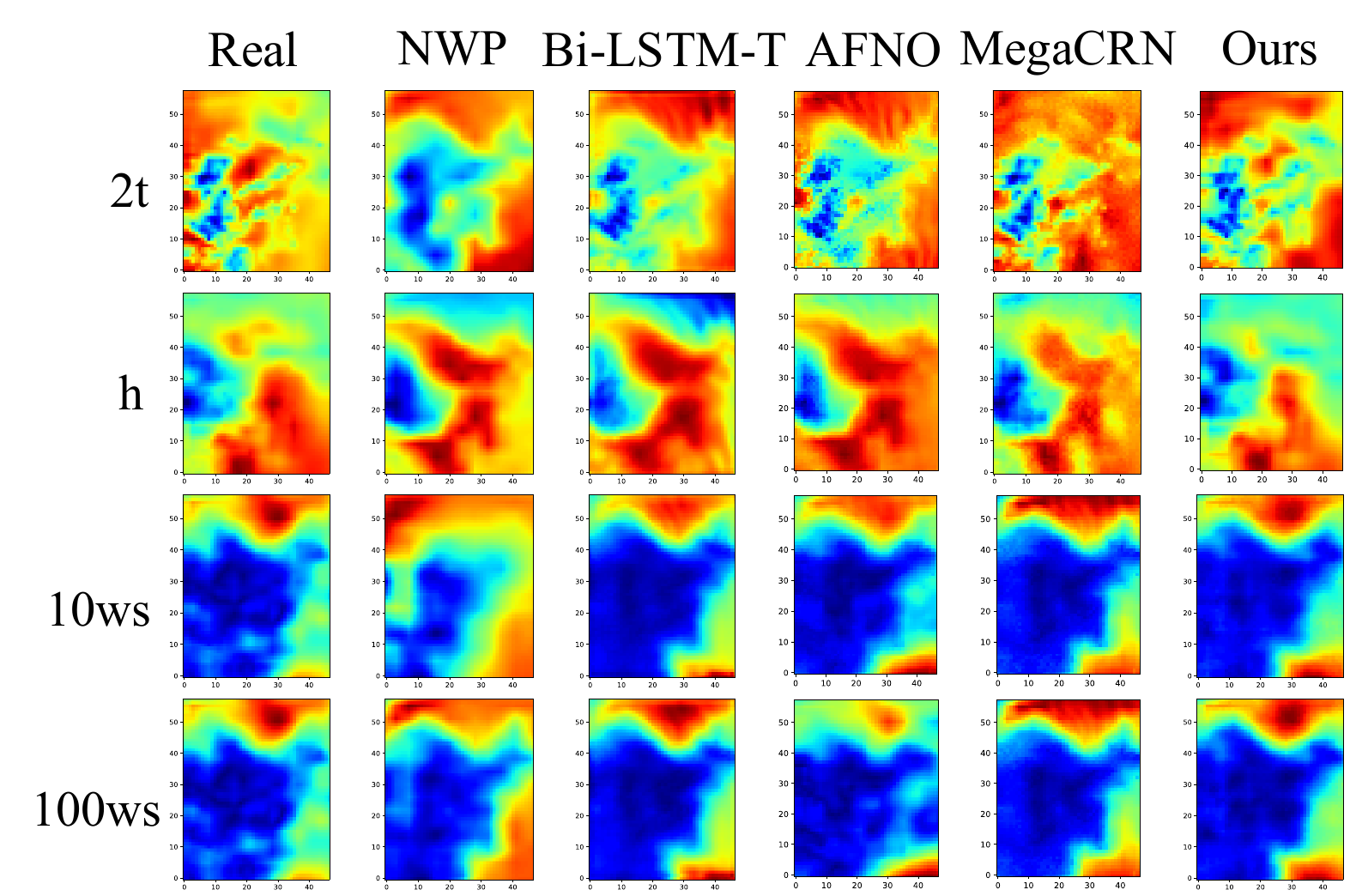}
    \caption{Visualization of Bias-Correction on Ningbo Dataset. Each row corresponds to a meteorological factor and each column corresponds to a method.}
    \label{fig:case1}
\end{figure}

\begin{figure}
    \centering
    \includegraphics[width=0.48\textwidth]{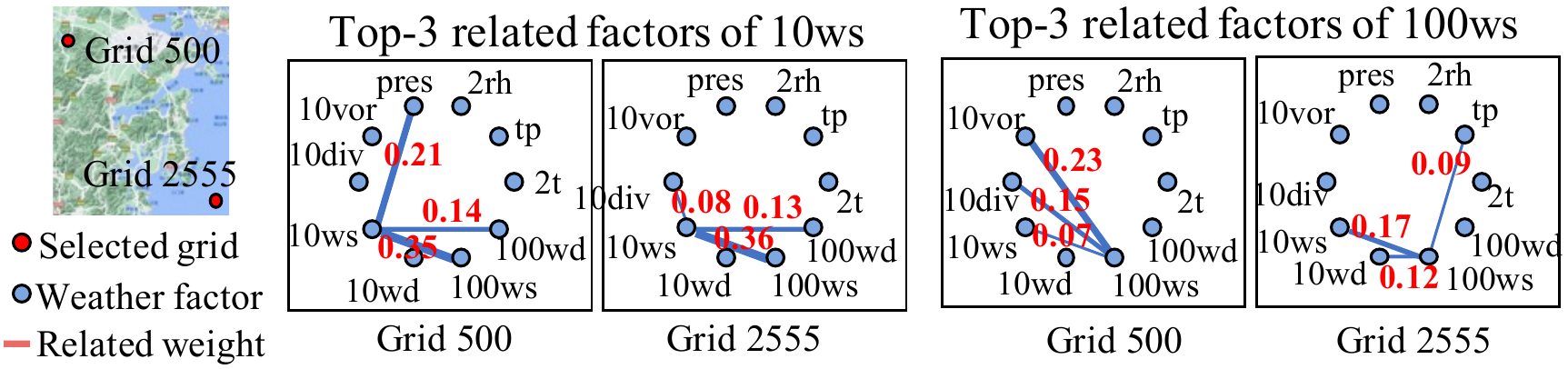}
    \caption{Illustration of Learned Factor Graphs of Two Grids on Two Different Types of Terrains.}
    \label{fig:factor_case}
\end{figure}

\noindent\textbf{Case study.} We visualize an example at 22:00 6/Apr/2021 about origin NWP and corrected results of baselines and WeatherGNN. As shown in Fig. \ref{fig:case1}, NWP has a large deviation from the ground truth, while baselines and WeatherGNN can effectively reduce this discrepancy. In addition, Fig. \ref{fig:factor_case} shows an example of the learned factor graphs selected randomly. Grid 500 locates in the mountains and Grid 2555 is in the ocean. 10ws and 100ws at both grids are closely related to each other. However, 10ws of Grid 500 is more relevant to pressure, as the mountainous regions have diverse terrain and are prone to rapid changes in air pressure, resulting in great effects on wind speeds. On the other hand, 10ws of Grid 2555 is affected by 10div (10m horizontal divergence, a measure of the local spreading or divergence of wind field in a horizontal plane at the height of 10 meters). One of the possible reasons is that on the sea horizon, the expansion of air in the horizontal direction has a significant impact on wind speed. More case studies about hierarchical structures, dynamic adjustment, and extreme weather are in Appendix. 

\section{Conclusion and Future Work}
In this paper, we propose WeatherGNN, a GNN-based model that leverages meteorological and spatial dependencies under the guidance of domain knowledge for local NWP bias-correction. Specifically, we introduce a factor GNN and a fast hierarchical GNN to capture area-specific meteorological dependencies adaptively and dynamic spatial dependencies efficiently, respectively. Extensive experimental results demonstrate the superiority of WeatherGNN. In the future, we will investigate WeatherGNN for other applications, e.g., weather forecasting and downsampling. We also plan to apply WeatherGNN to larger real-world regions to investigate its scalability and robustness.

\section*{Contribution Statement}
Binqing Wu and Weiqi Chen contributed equally to this work as co-first authors. Liang Sun and Ling Chen are the corresponding authors.

\bibliographystyle{named}
\bibliography{ijcai24}

\end{document}